\documentclass[acmsmall,nonacm]{acmart}

\renewcommand\footnotetextcopyrightpermission[1]{} 
\usepackage{booktabs}
\usepackage{amssymb}
\usepackage{xcolor}
\usepackage{graphicx}
\usepackage{amsmath}
\usepackage{subfig}
\usepackage{glossaries}

\usepackage[normalem]{ulem}

\begin{document}

\title{City-GAN: Learning architectural styles using a custom Conditional GAN architecture}

%
%
%
%
%

\author{Maximilian Bachl}
\email{maximilian.bachl@gmail.com}

\author{Daniel C. Ferreira}
\email{daniel.ferreira.1@gmail.com}

\renewcommand{\shortauthors}{Bachl and Ferreira}

\begin{abstract}
Generative Adversarial Networks (GANs) are a well-known technique that is trained on samples (e.g.~pictures of fruits) and which after training is able to generate realistic new samples. Conditional GANs (CGANs) additionally provide label information for subclasses (e.g.~apple, orange, pear) which enables the GAN to learn more easily and increase the quality of its output samples. We use GANs to learn architectural features of major cities and to generate images of buildings which do not exist. We show that currently available GAN and CGAN architectures are unsuited for this task and propose a custom architecture and demonstrate that our architecture has superior performance for this task and verify its capabilities with extensive experiments. 
\end{abstract}

\maketitle




\newacronym{gan}{GAN}{Generative Adversarial Network}
\newacronym{cgan}{CGAN}{Conditional Generative Adversarial Network}
\newacronym{cnn}{CNN}{Convolutional Neural Network}

\section{Introduction}

\glspl{gan} are a neural network architecture whose main purpose is generating previously unseen instances of data following a previously learned distribution. For example, they can be used to generate new handwritten digits or faces of people who do not exist \cite{goodfellow_generative_2014}. 

This works by having two separate neural networks, the generator and the discriminator. The generator's task is to create realistic samples and the discriminator's task is to determine whether a given sample is real or a fake from the generator. The input to the generator is a vector of noise, and its output a data sample. The discriminator's input is a data sample and its output is a decision whether the sample if real or fake. Over the course of training a \gls{gan}, the generator learns to create more realistic samples while the discriminator gets better at distinguishing real and fake instances. 

\glspl{cgan} \cite{mirza_conditional_2014} were introduced recently after \glspl{gan}. Their modification is an additional label input for both the generator and the discriminator. For example, for learning to generate digits, the labels would be digits from 0 to 9. For example, if one gave the label ``2'' to the generator it would try to generate a digit that resembles ``2''. The discriminator then would also get the label ``2'' so that it knows that the sample it is looking at should be ``2'' no matter if it is real or fake. 

\glspl{cgan} can be used for tagging images by utilizing the discriminator \cite{mirza_conditional_2014}. It is also possible to use the labels in the generator to transition between different cases \cite{gauthier_conditional_2015}, like for example transitioning from ``1'' to ``7''. 

Commonly, \glspl{cgan} are combined with \glspl{cnn} to be able to process image data \citep{radford_unsupervised_2015}. For the generator this is done by concatenating the label with the noise vector that is fed into the generator. For the discriminator the image is processed by a regular \gls{cnn}. After the \gls{cnn} there is one (or more than one) fully connected layer, which takes as its input the output of the \gls{cnn} as well as the label. Finally the discriminator outputs its decision regarding the realness of the sample. 

We present one new dataset for \glspl{gan}: The dataset consists of thousands of images of facades of buildings in various cities. Each image also has a label, which indicates the city of the building depicted. 

We show that both regular Convolutional \glspl{gan} as well as Convolutional \glspl{cgan} have significant shortcomings for our use case. We then propose a new custom architecture, which delivers higher-quality results as well as more stable convergence for our datasets. 

In summary, we make the following contributions:
\begin{itemize}
\item a new neural network architecture for Convolutional \glspl{cgan}
\item two datasets of facades
\item insight about the architectural features of the cities under study
\item program code that allows practitioners to reproduce our experiments and generate similar datasets
\end{itemize}

All our code and the dataset we used are available at \url{https://github.com/muxamilian/city-gan}.

\section{Overview}

We implemented \textit{City-GAN}, a \textit{Generative Adversarial Network} \citep{goodfellow_generative_2014} that can generate facades of buildings in cities it was trained on. 

To achieve this, we followed these steps: 
\begin{enumerate}
\item We chose four cities for our analysis: Vienna, Paris, Amsterdam and Manhattan.
\item We created a list of valid addresses in these cities.
\item We downloaded images from Google Street View from a randomly sampled subset of these addresses in each city. We chose the altitude angle of the camera to be 20 degrees. 
\item We filtered images which fulfilled the following criteria:
\begin{itemize}
\item It is possible to find a way from the left to the right edge of the image that only goes over facades and has no gaps in between.
\item The building does not show \textit{anomalous structures}. For example, we consider construction sites as anomalous because they do not represent the typical cityscape. 

\end{itemize}
\begin{figure}[h]
\includegraphics[width=0.49\columnwidth{}]{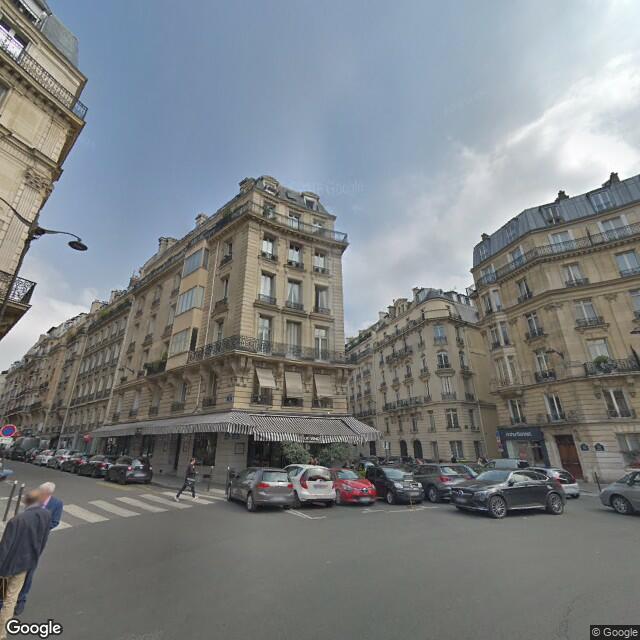}%
\hspace{0.015\columnwidth}%
\includegraphics[width=0.49\columnwidth{}]{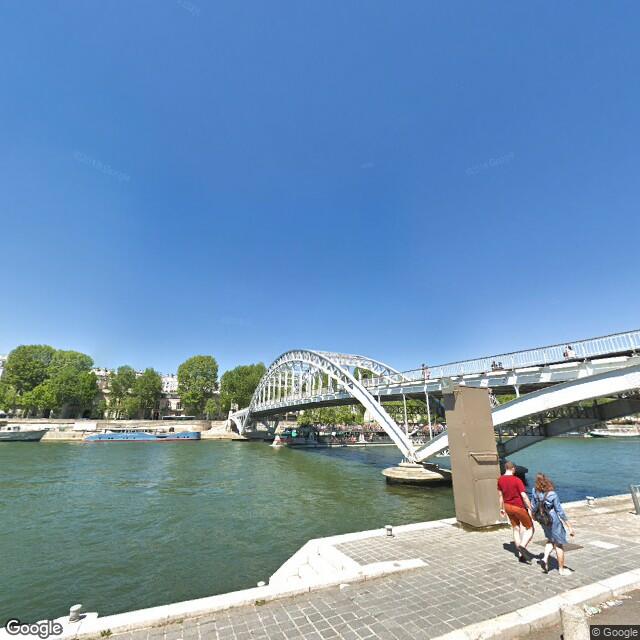}
\caption{Two examples of pictures that we did not include in our dataset: on the left image there are gaps in the facades, the right one is anomalous (not a facade).}
\label{fig:example_invalid}
\end{figure}

\item After filtering we had 1000 images per city (4000 in total). 
\item We experimented with different types of GANs and different architectures.
\item We used another, larger dataset \citep{zamir_repo_2019} and checked whether a significantly higher number of samples improves the learning performance. This dataset consists of 500,000 facades each for Amsterdam, Washington DC, Florence, Las Vegas and Manhattan. These images are, however, not curated. The only restriction we made is that the altitude angle must be between 0 and 40 degrees. Otherwise, these images are randomly sampled from the cities with no further restrictions. 
\end{enumerate}

We tried different GAN architectures during the course of our experimentation. For all our experiments we used the following data augmentation methods: horizontal flipping with probability 0.5 as well as random cropping of approximately 85\% of the original width and height of the images. We experimented with different cropped image sizes: 64, 128 and 256 pixels in height and width. For a final image of 256 pixels, we would input an image of 300 pixels and randomly crop it to 256.

\section{Regular GAN}

First, we used a regular deep convolutional GAN \citep{radford_unsupervised_2015} and didn't distinguish between the different cities. The code we used we got from the official PyTorch examples \citep{pytorch_deep_2019} which implements the procedure outlined by \citet{radford_unsupervised_2015}. However, from visually inspecting the results, we concluded that the GAN has a hard time learning the data distribution. 

The generator starts with a vector of multivariate gaussian noise (length 100) and continuously deconvolutes it (kernel size 4, stride 2) until it outputs the image of the final size. 

The discriminator gets the image and continuously convolutes it (kernel size 4, stride 2) until it is of dimension $1\times 1 \times n_\text{channels}$. Then there is one linear layer followed by a sigmoid function that outputs whether the sample is real or fake. 

\begin{figure}[h]
\includegraphics[width=\columnwidth]{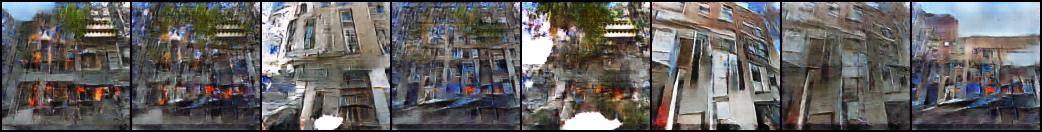}
\caption{Eight random samples from the generator of the regular GAN.}
\label{fig:gan}
\end{figure}

\autoref{fig:gan} shows the results of our generator after sufficient training iterations. We see that while its output looks like facades, the output is clearly faulty and can be easily characterized by a human as being fake. 

\section{Conditional GAN 1}

After realizing that a regular GAN has problems learning the data distribution we used a conditional GAN (CGAN) \citep{mirza_conditional_2014}. The rationale behind this was that if we tell the GAN which city a picture is from, it can focus on learning facades and doesn't get confused by different cities' styles. 
The difference between a GAN and a CGAN is that the latter receives label information both in the generator and discriminator.
As such, the discriminator has more information to distinguish between real and fake pictures, and the generator has an incentive to develop features specifically for each label.

We implemented it ourselves based on the code of the regular GAN. The generator of the CGAN gets a one-hot-encoded vector of the cities alongside the multivariate Gaussian distribution and can thus consider the additional information about the city that the discriminator expects in the image generation process. For instance, when the generator is supposed to output an image of Amsterdam, it gets the vector $\begin{pmatrix}1, 0, 0, 0 \end{pmatrix}^\top$ attached to its input. Thus the input to the generator network is a vector of length 104 with our four cities (100+4). 

The discriminator of the GAN proceeds as in the regular GAN and uses a convolutional neural network to process the input image. After the convolutional layers we concatenate their output with the one-hot-encoded vector of the city and pass it through a dense layer followed by a ReLU function. After that we have another dense layer followed by a sigmoid function that outputs whether the sample is real or fake.  

\begin{figure}[h]
\includegraphics[width=\columnwidth]{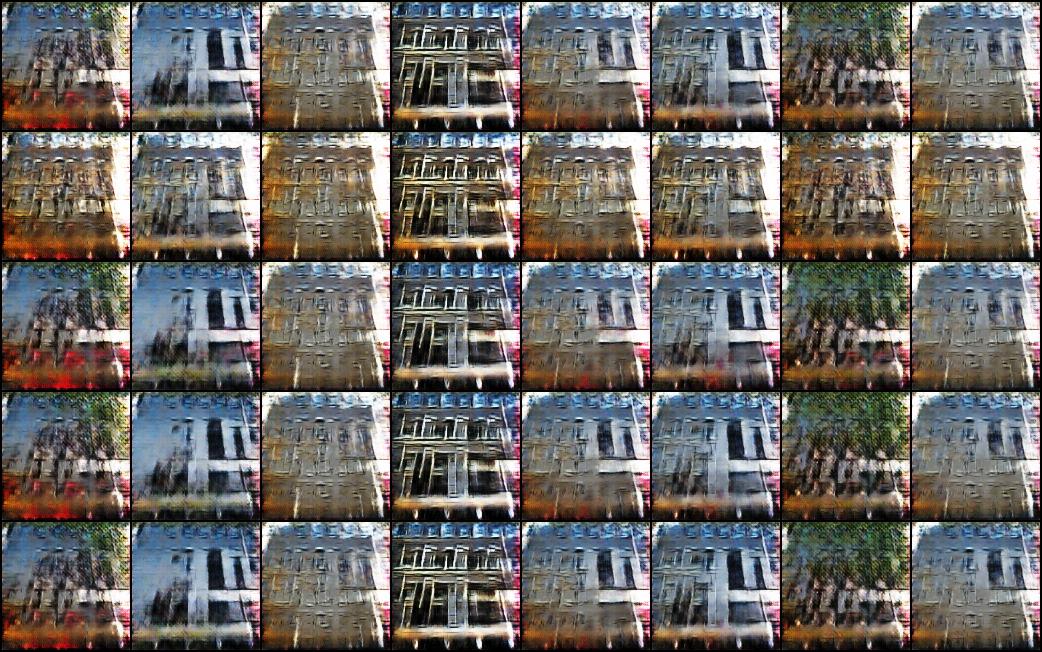}
\caption{Our first attempt at a CGAN. Each column represents one random input to the generator. The rows represent: the average of all cities, Amsterdam, Manhattan, Paris and Vienna. }
\label{fig:daniel_net}
\end{figure}

\autoref{fig:daniel_net} shows that the results are even worse than for the regular GAN.

\section{Conditional GAN with custom architecture} 

\begin{figure}
\includegraphics[width=\columnwidth]{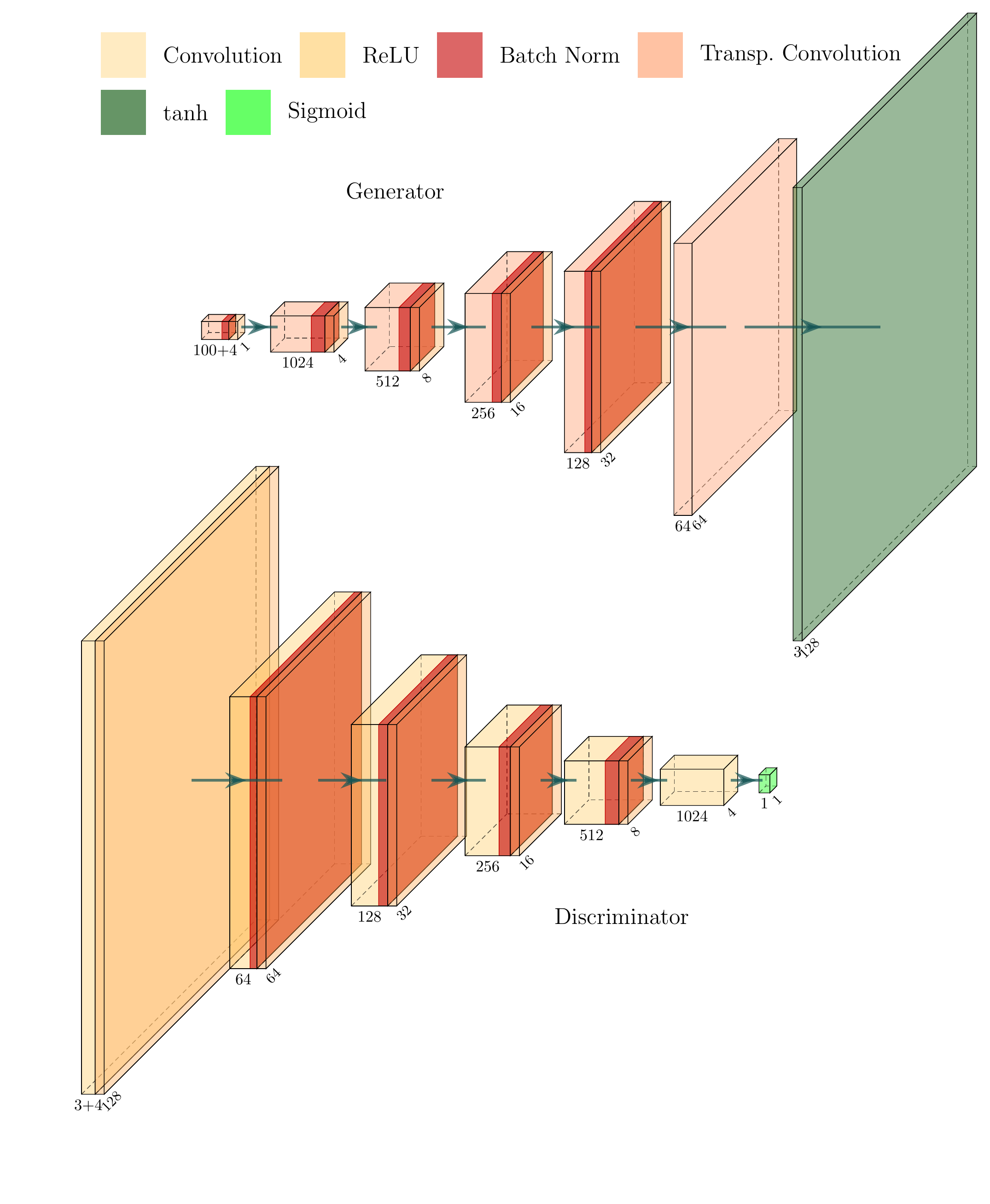}
\caption{The architecture of our neural network is inspired by \cite{pytorch_deep_2019}. The difference in our custom architecture is that labels are fed into the generator (top) as well as into the discriminator (bottom) alongside the data.}  
\label{fig:maxnet}
\end{figure}

After achieving suboptimal results with this CGAN architecture we tried a different approach and modified the discriminator: instead of adding the one-hot-encoded vector with the city information after the convolution, we decided to add it as an input to each pixel in the beginning of the convolution. Thus, additional to the RGB values, each pixel also has the one-hot-encoded information of the city. For example, with four cities, each pixel has 7 color channels: 3 for the colors and 4 for the one-hot-encoded cities (\autoref{fig:maxnet}). 

While this approach sounds wasteful because of the duplicate information that is fed into the convolutional layer, it achieved significantly better results. One possible explanation for this improvement is that by adding the city information to the input picture, we help the kernels of the convolution to work differently for each city. 

\begin{figure}[h]
\includegraphics[width=\columnwidth]{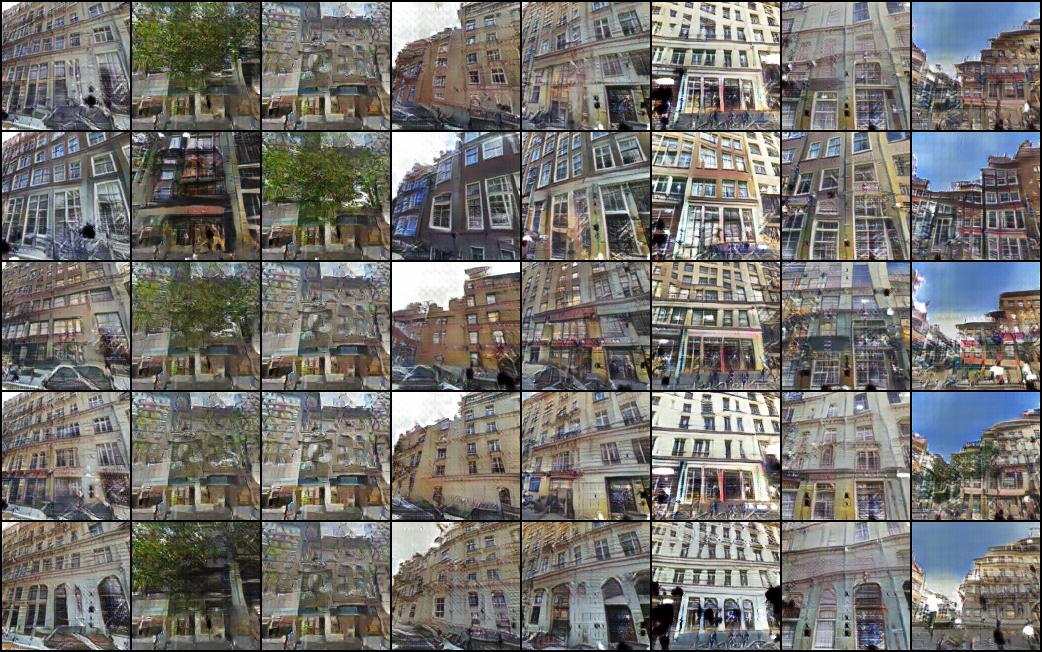}
\caption{Our custom CGAN architecture. Each column represents one random input to the generator. The rows represent: the average of all cities, Amsterdam, Manhattan, Paris and Vienna.}
\label{fig:hack}
\end{figure}

\autoref{fig:hack} shows that the results of our custom CGAN are significantly better than with our previous attempt and are also better than the normal GAN. 

\section{Influence of image size on the quality of generated images}

\begin{figure}[h]
\centering
\includegraphics[width=0.75\linewidth]{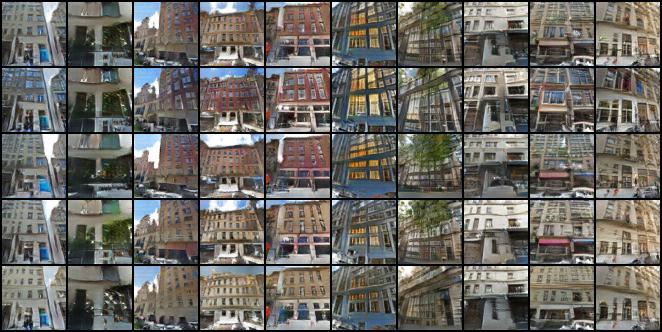}
\caption{Our custom CGAN trained on images of size $64\times 64$. Each column represents one random input to the generator. The rows represent: the average of all cities, Amsterdam, Manhattan, Paris and Vienna.}
\label{fig:smaller}
\end{figure}

We also tested the effect of image size on the quality of the generated images. For this, we used images with a length and width of 64, 128 and 256 pixels. Our general conclusion is that smaller image sizes improve quality: with 64 pixels \autoref{fig:smaller}, the generated facades looked good and had no or only very minor flaws. With 128 we got a larger number of artifacts and faulty images but we deemed quality to be still acceptable. With 256, we couldn't get satisfying results any longer: the generator never converged and changed its output over time when fed the same random input. Furthermore, the generator did not generalize over different cities and, when inputting an average of all cities using the input vector $\begin{pmatrix}\frac{1}{4}, \frac{1}{4}, \frac{1}{4}, \frac{1}{4}\end{pmatrix}^\top$---which gives equal importance to all cities---the obtained image was only noise. Thus, the generator only learned to output facades for each city but did not learn which features all cities have in common. During the course of our further experiments we only used images of resolution $128\times 128$.

\section{Using a larger, uncurated dataset}

Finally, we wanted to investigate whether a larger dataset could significantly improve the quality of generated images. For this we used a dataset of Stanford University \citep{zamir_repo_2019} and extracted 500,000 images for the cities of Amsterdam, Washington DC, Florence, Las Vegas and Manhattan. These images are not curated, thus it is not guaranteed that they actually show facades. 


We noticed that Washington DC and Las Vegas are especially problematic for our neural network. 
In Washington DC, the majority of images depict trees, and in Las Vegas there are few multi-story buildings. 

Thus we also tried training the CGAN with only these three cities, hoping this would increase the quality of learned images.

\begin{figure}[h]
\includegraphics[width=\columnwidth]{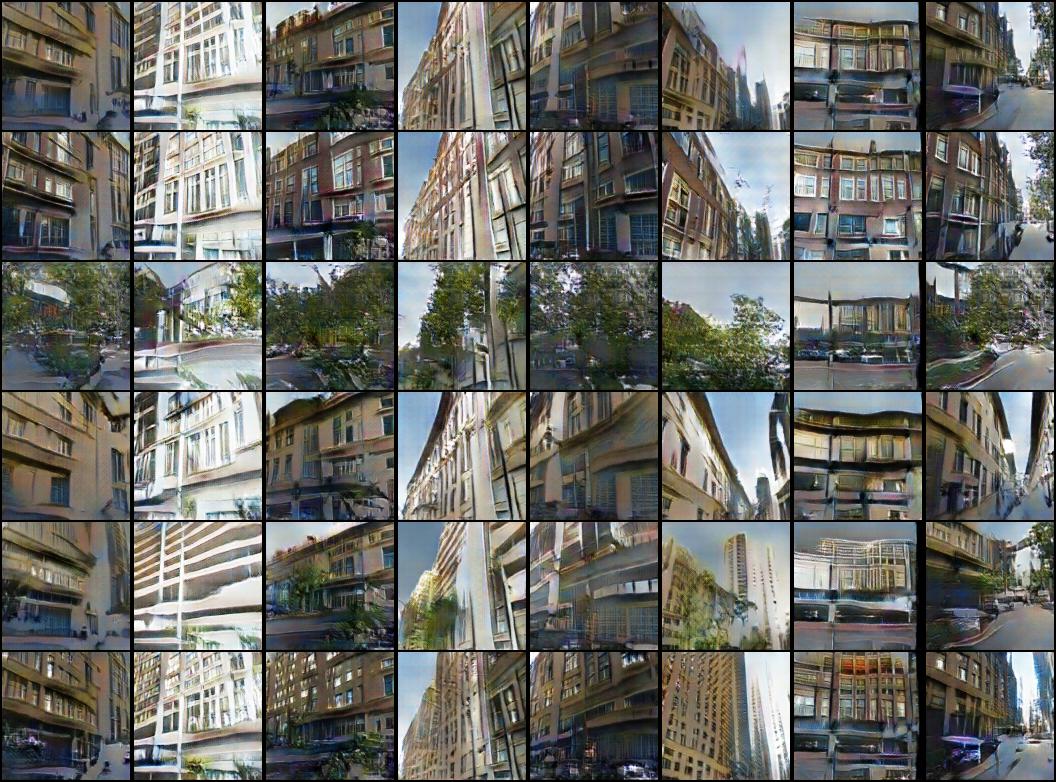}
\caption{Our custom CGAN trained on the Stanford dataset. Each column represents one random input to the generator. The rows represent: the average of all cities, Amsterdam, Washington D.C., Florence, Las Vegas and Manhattan.}
\label{fig:stanford_net}
\end{figure}

Generally speaking we think that the quality of the generated images is the same as the one of our own dataset \autoref{fig:stanford_net}. However, while with our dataset we only learn facades, with the Stanford dataset we also learn entire streets with building on the left and right side of the street. 

\section{Traveling Between Cities}

\begin{figure}[h]
	\centering
	\subfloat[Amsterdam$\to$Florence]{\includegraphics[width = 0.49\linewidth]{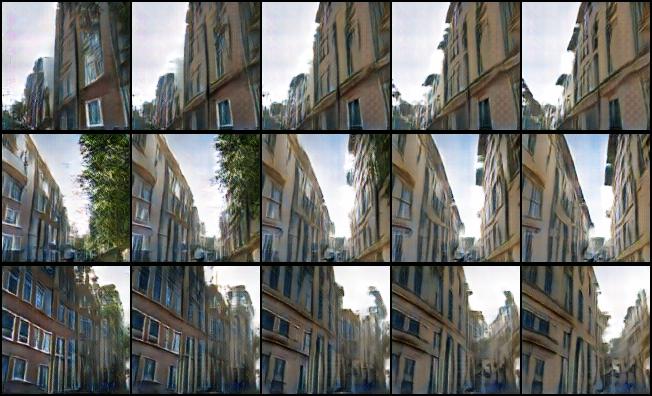}\label{fig:ta}}
	\hspace{0.01\linewidth}
	\subfloat[Amsterdam$\to$Washington DC]{\includegraphics[width = 0.49\linewidth]{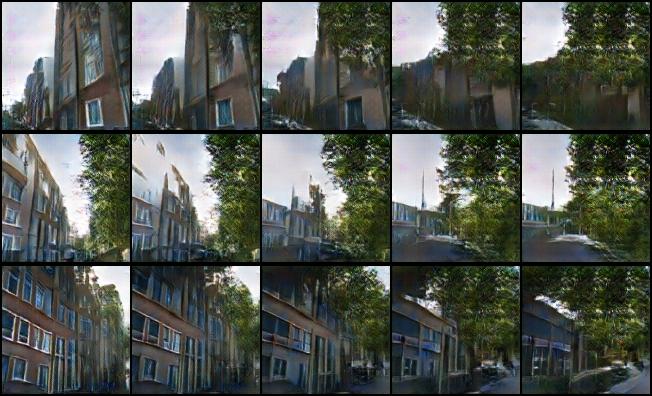}\label{fig:tb}}\\
	\subfloat[Amsterdam$\to$Manhattan]{\includegraphics[width = 0.49\linewidth]{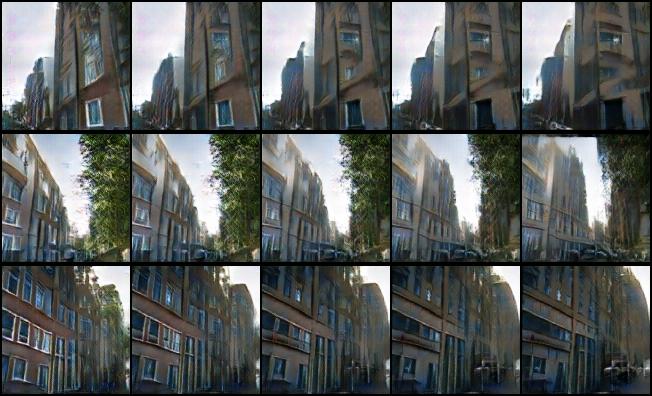}\label{fig:tc}}
	\hspace{0.01\linewidth}
	\subfloat[Europe$\to$US]{\includegraphics[width = 0.49\linewidth]{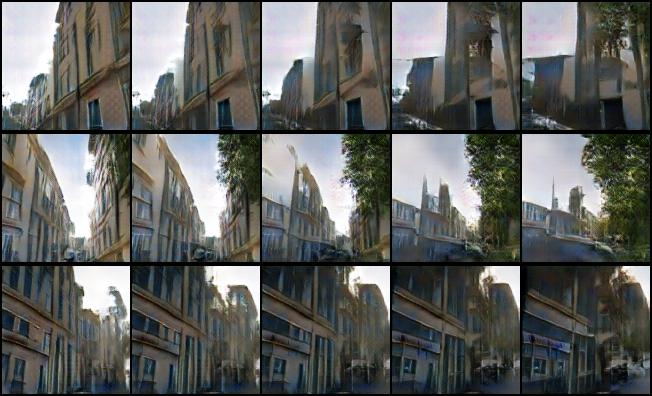}\label{fig:td}} 
	\caption{Outputs of our CGAN when interpolating between locations.
	Each row contains one cityscape, interpolated across the locations specified in the subtitle.
}
	\label{fig:transfer}
\end{figure}

To understand if our network learns characteristics of each city, we can compare the output of the generator for the same vector with different labels.
We show a 5-step interpolation for 3 distinct random vectors in Figure~\ref{fig:transfer}.
We show results interpolating from Amsterdam to each one of Florence, Washington DC and Manhattan.
Additionally, we show also results of interpolating from ``Europe'' (half Amsterdam/half Florence) to the ``US'' (a third of each American city).
These results are from the CGAN with the uncurated dataset described in the last section.

The first thing we notice is that the color palette of the cities is quite different.
For example, Amsterdam is quite redish, while Florence has tones closer to beige (\autoref{fig:ta}).
Another obvious characteristic is that Washington DC has a lot of trees (\autoref{fig:tb}).
In all the examples with it, the amount of trees greatly increases when compared to Amsterdam.

There are also some more nuanced changes that characterize the stylistic architectural differences between cities.
A good example of this is show between Amsterdam and Florence: distinct roofs, with breaks between them can be seen.
This is most noticeable in the second row of \autoref{fig:ta}.
Between Amsterdam and Manhattan there are also two strong distinctions: Manhattan has taller buildings, with more smaller windows.
In the second row of Figure~\ref{fig:tc}, the buildings grow as the label approaches Manhattan, and in the third row the windows split into smaller windows.

\begin{figure}[h]
	\centering
	\includegraphics[width=0.49\linewidth]{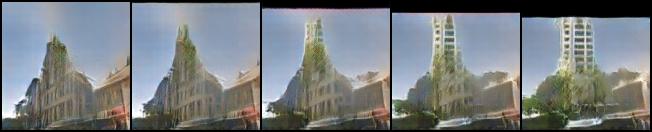}
	\caption{Transition from a tall European church to an American skyscraper.}
	\label{fig:skyscraperchurch}
\end{figure}

We also show the transition between Europe and the US.
The US images look quite unrealistic, as they are an average between 3 cities with quite different styles.
However, one can identify an increase of trees (Washington DC has really a lot of trees) and different angles of photography (buildings are farther from the camera in Washington DC and Las Vegas).
Both of these differences are evident in the second row of \autoref{fig:td}.
Another good example of the Europe/US relation is in \autoref{fig:skyscraperchurch}, which shows an European style church being transformed into an American style skyscraper.

\begin{figure}[h]
	\centering
	\includegraphics[width=0.49\linewidth]{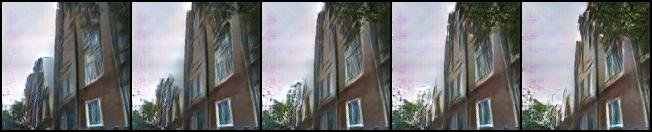}
	\caption{Transition from Amsterdam to Amsterdam-Manhattan (This means that we set the one-hot vector with 1 for Amsterdam and -1 for Manhattan).}
	\label{fig:minusmanhattan}
\end{figure}

Another thing we can try is to experiment with subtracting stylistic features.
We can for example subtract Manhattan's label from Amsterdam's, and intuitively that should remove typical Manhattan features from the Amsterdam image.
Indeed, we do just that in \autoref{fig:minusmanhattan}, and observe that the modern looking building in the background in Amsterdam completely disappears, after subtracting Manhattan. Also, if you subtract Manhattan, more trees appear. Apparently, in Manhattan suppresses the output of trees because there are not many there. 

\begin{figure}[h]
	\centering
	\includegraphics[width=0.7\linewidth]{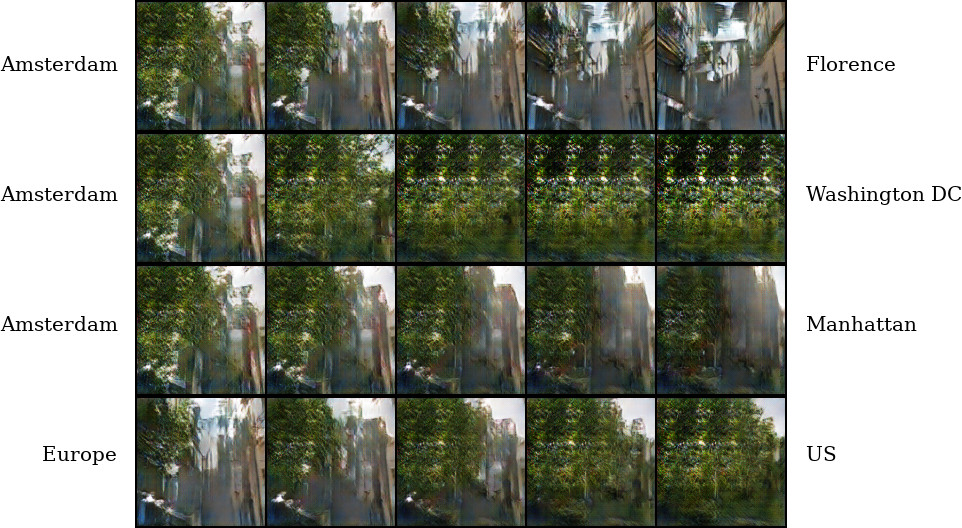}
	\caption{Transition from Amsterdam to each of Florence, Washington DC, and Manhattan (first 3 rows), and from Europe to the US (last row).}
	\label{fig:bad}
\end{figure}

Not all images obtained are realistic, and some are even quite bad.
In Figure~\ref{fig:bad}, we show some of these bad examples.
The transitions are in the same order as in \autoref{fig:transfer}, but with only one fixed random vector instead of 3.

\section{Conclusion}

In this paper we sought to generate images of facades containing the stylistic features of specific cities.
We experimented with multiple GAN architectures, and propose a custom architecture.
Our custom architecture produced more realistic images than the alternative architectures, and was capable of generating architectural features specific to each city.
By using a continuous label space, we were also able to transition one facade between cities, or do averages over them.

The results suggest that our architecture is also better suited for other tasks than the existing architectures.
Investigating its effects on other tasks is a feasible future research path.

We have made all our code and used data available, so that the community can both reproduce and improve on our results.

\section*{Acknowledgement}
The Titan Xp used for this research was donated by the NVIDIA Corporation.

\bibliographystyle{ACM-Reference-Format}
\bibliography{biblio}

\end{document}